\documentclass[conference]{IEEEtran}
\IEEEoverridecommandlockouts
\usepackage{cite}
\usepackage{amsmath,amssymb,amsfonts}
\usepackage{algorithmic}
\usepackage{graphicx}
\usepackage{textcomp}
\usepackage{xcolor}
\usepackage{svg}
\usepackage{multirow}
\usepackage{booktabs}
\usepackage{url}

\def\BibTeX{{\rm B\kern-.05em{\sc i\kern-.025em b}\kern-.08em
    T\kern-.1667em\lower.7ex\hbox{E}\kern-.125emX}}
\begin{document}

\title{Promptable Fire Segmentation: Unleashing SAM2's Potential for Real-Time Mobile Deployment with Strategic Bounding Box Guidance
}

\author{\IEEEauthorblockN{1\textsuperscript{st} Emmanuel U. Ugwu}
\IEEEauthorblockA{\textit{School of Computer Science and Technology} \\
\textit{University of Science and Technology of China}\\
Hefei, Anhui, China \\
ugwuemmanuelugo2000@mail.ustc.edu.cn}
~\\

\and
\IEEEauthorblockN{2\textsuperscript{nd} Zhang Xinming*}
\IEEEauthorblockA{\textit{School of Computer Science and Technology} \\
\textit{University of Science and Technology of China}\\
Hefei, Anhui, China \\
xinming@ustc.edu.cn}
*Corresponding author
}

\maketitle

\begin{abstract}
Fire segmentation remains a critical challenge in computer vision due to flames' irregular boundaries, translucent edges, and highly variable intensities. While the Segment Anything Models (SAM and SAM2) have demonstrated impressive cross-domain generalization capabilities, their effectiveness in fire segmentation—particularly under mobile deployment constraints—remains largely unexplored. This paper presents the first comprehensive evaluation of SAM2 variants for fire segmentation, focusing on bounding box prompting strategies to enhance deployment feasibility. We systematically evaluate four SAM2.1 variants (tiny, small, base\_plus, large) alongside mobile-oriented variants (TinySAM, MobileSAM) across three fire datasets using multiple prompting strategies: automatic, single positive point (SP), single positive point + single negative point  (SP+SN), multiple positive points (MP), bounding box (Box), and hybrid variants (Box+SP and Box+MP). Our experimental results demonstrate that bounding box prompts consistently outperform automatic and single point-based approaches, with Box+MP achieving the highest mean IoU (0.64) and Dice coefficient (0.75) on the Khan dataset. Lightweight variants such as TinySAM and MobileSAM further reduce memory and computational costs, making them more suitable for latency-tolerant edge scenarios. Overall, this work provides critical insights for deploying promptable segmentation models in fire monitoring systems and establishes benchmarks for future research in domain-specific SAM applications. Code is available at: \url{https://github.com/UEmmanuel5/ProFSAM}.

\end{abstract}


\section{Introduction}
\label{sec:intro}

Fire incidents continue to pose significant threats to human lives, critical infrastructure, and ecological systems worldwide. Traditional fire detection systems rely on smoke detectors and thermal sensors, which provide limited spatial information and often suffer from delayed response times or false alarms in complex environments \cite{ahrens2010smoke}. Computer vision-based fire detection offers superior spatial awareness through pixel-level segmentation, enabling precise localization and extent estimation of fire regions.

However, fire segmentation presents unique computational challenges. Flames exhibit irregular, amorphous shapes with translucent boundaries that vary dramatically under changing illumination conditions. The dynamic nature of fire, combined with smoke occlusion and background clutter, makes traditional segmentation approaches particularly challenging \cite{khan2023efficient}. While deep learning models such as U-Net variants have shown promise, they typically require extensive task-specific training and may not generalize well across different fire scenarios \cite{muhammad2023efficient}.

The recent introduction of Meta AI's Segment Anything Model (SAM) \cite{kirillov2023segment} and its successor SAM2 \cite{ravi2024sam2} represents a paradigm shift toward universal, promptable segmentation. These models demonstrate remarkable zero-shot generalization across diverse domains using simple prompts (points, boxes, or masks) without requiring task-specific training. SAM2 extends this capability to video sequences through memory-enabled temporal consistency, making it particularly relevant for real-time fire monitoring applications.

Despite SAM's success in medical imaging \cite{senguptamedical,yurobotics,zhangunleashing} and camouflage object detection \cite{lian2024evaluation,tang2024evaluating}, its application to fire segmentation remains largely unexplored. Recent studies have begun investigating SAM for wildfire applications: Marinaccio and Afghah \cite{marinaccio2025seeing} utilized SAM for wildfire temperature inference, while Pesonen et al. \cite{pesonen2025detecting} employed SAM-enhanced pseudo labels for UAV-based wildfire detection. However, these works primarily focus on specific applications rather than comprehensive evaluation of SAM's fire segmentation capabilities.

Moreover, the deployment of SAM on mobile and edge devices presents additional challenges. While mobile variants such as MobileSAM, TinySAM, and NanoSAM have been developed to address computational constraints \cite{sun2024efficient}, their performance in fire segmentation scenarios has not been systematically evaluated. This gap is particularly critical given the importance of real-time fire detection in emergency response scenarios.

\begin{figure}[htbp]
    \centering
    \includegraphics[width=0.95\linewidth]{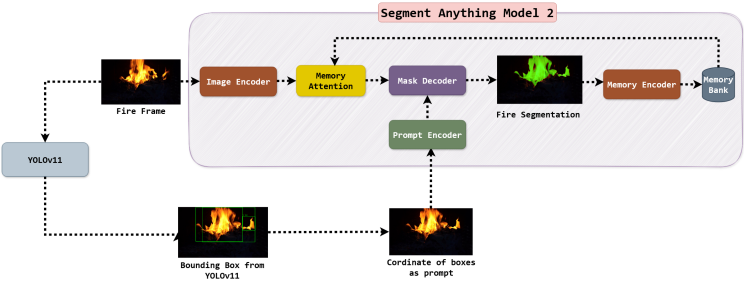}
    \caption{Fire segmentation pipeline using SAM2. An input fire video frame is processed with bounding box prompts through the SAM2 model, producing a temporally consistent segmentation mask. The output is visualized as a green fire overlay on the original frame for real-time monitoring applications.}
    \label{fig:proposed_pipeline}
\end{figure}

This work addresses these limitations by conducting the first systematic evaluation of SAM2 variants for fire segmentation, with particular emphasis on prompt engineering strategies and mobile deployment feasibility (Figure~\ref{fig:proposed_pipeline}). We demonstrate that strategic bounding box prompting significantly enhances segmentation performance while enabling practical deployment on resource-constrained devices. Our comprehensive analysis across multiple datasets provides crucial insights for developing next-generation fire monitoring systems that leverage foundation models' generalization capabilities while meeting real-world deployment constraints.

\section{Related Work}
\label{sec:related}

In the field of image segmentation, several techniques such as fuzzy histogram thresholding \cite{thanammal2014effective}, fuzzy C-means with watershed integration \cite{kaur2014fcm}, and fuzzy color segmentation via watershed transforms \cite{richard2013fuzzy} have been explored to address challenges in delineating complex image regions. In parallel, fire segmentation research primarily relied on color-based heuristics in RGB or HSV color spaces, often combined with motion analysis and hand-crafted feature extractors \cite{toulouse2015automatic}. While these approaches provided simple and interpretable solutions, they were highly sensitive to illumination changes and fire-like objects, leading to frequent false positives. To address these challenges, subsequent studies incorporated fuzzy logic and machine learning techniques \cite{sowahfuzzy, sowah2018neurofuzzy}, which improved robustness to uncertainty but still depended heavily on manually designed rules.

The advent of deep learning significantly improved fire detection, with convolutional neural networks (CNNs) achieving superior performance in both classification and localization tasks \cite{khan2024yolo}. For fire segmentation, U-Net-based architectures with lightweight encoders such as MobileNet and ShuffleNet have gained popularity due to their computational efficiency. For example, Khan et al. introduced an efficient U-Net variant that achieved competitive segmentation performance with only 1.49 MB of parameters by leveraging ShuffleNet as the encoder for lightweight feature extraction \cite{khan2023efficient, muhammad2023efficient}. However, these deep learning models still require extensive supervised training on fire-specific datasets and often fail to generalize well to unseen environments without retraining. This limitation has motivated the exploration of foundation models like the Segment Anything Model (SAM), which can adapt to new domains through prompting rather than retraining.

SAM \cite{kirillov2023segment} introduced promptable segmentation, leveraging over 1.1 billion training masks, to achieve remarkable zero-shot generalization across diverse domains. Its architecture consists of a Vision Transformer (ViT) image encoder, a prompt encoder capable of processing points, boxes, and masks, and a lightweight mask decoder. By design, SAM can generate multiple valid masks per prompt to address ambiguity and achieves sub-50ms interactive inference. Its performance scales with encoder capacity: larger backbones such as ViT-H deliver higher quality on fine-grained tasks but at significant computational cost, while smaller encoders reduce inference time with some accuracy trade-offs. Extending SAM to video, SAM2 \cite{ravi2024sam2} introduces a memory-enabled transformer architecture with temporal attention, enabling consistent object tracking across frames from sparse prompts. This mechanism makes SAM2 more suitable for dynamic fire monitoring scenarios. 

Recent studies have started exploring SAM's fire segmentation potential. Liu et al. \cite{liusafe2025} employed SAM for accurate and complete burned area extraction, while Marinaccio and Afghah \cite{marinaccio2025seeing} integrated SAM for wildfire temperature inference from RGB imagery. However, these studies target specific application scenarios rather than systematically evaluating SAM's overall performance for fire segmentation tasks.

One major challenge with deploying SAM in real-world fire monitoring systems is its computational demand, which limits usage on mobile and edge devices. To address this, several efficient variants have been developed \cite{sun2024efficient}. MobileSAM \cite{zhao2023mobilesam} reduces parameters via a decoupled knowledge distillation strategy that transfers representations from the heavy ViT-H encoder to a compact TinyViT encoder, thereby maintaining mask decoder compatibility while achieving up to 60× parameter reduction and near real-time inference. Building further, TinySAM \cite{shu2024tinysam} employs full-stage distillation with mask-weighted losses, quantization, and a hierarchical ``segment everything'' strategy, which halves dense proposal inference time with minimal accuracy loss. NanoSAM \cite{nanosam} distills MobileSAM to a ResNet-18 backbone and compiles it with TensorRT for NVIDIA Jetson deployment, reaching real-time segmentation on embedded hardware. More recently, Bonazzi et al. \cite{bonazzi2025picosam2} introduced PicoSAM2, which achieves real-time segmentation on the Sony IMX500 edge AI sensor. These variants exemplify how model mechanisms— whether encoder distillation, quantization, or hardware-aware deployment pipelines— systematically drive efficiency-performance trade-offs, underscoring their potential for latency-critical fire monitoring.

\begin{figure}[htbp]
    \centering
    \includegraphics[width=0.95\linewidth]{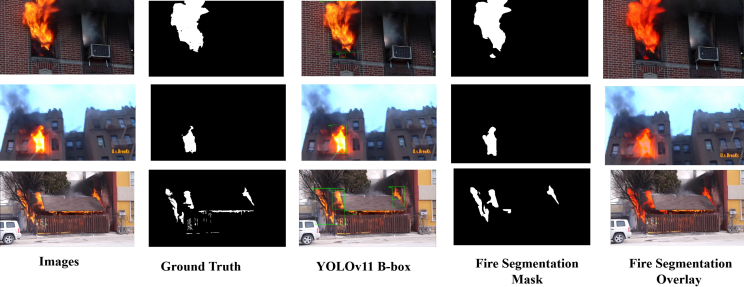}
    \caption{Step-by-step fire segmentation using SAM2. 
    From left to right: (1) original fire image, (2) ground truth mask, (3) fire region detection using YOLOv11, (4) SAM2-generated mask using the YOLOv11 bounding box as prompt, and (5) final fire segmentation visualized as a red overlay on the original image.}
    \label{fig:bbox_prompt_pipeline}
\end{figure}

Another active research area focuses on improving SAM's segmentation performance through prompt engineering. Rafaeli et al. \cite{rafaeli2024prompt} showed that bounding box prompts yield superior performance under varying lighting conditions and image resolutions, while Pei et al. \cite{pei2024evaluation} reported that SAM2's segmentation quality significantly degrades when box prompts are omitted. Building on this, Li et al. \cite{li2024amsam} proposed AM-SAM, which incorporates automated prompting and mask calibration to improve segmentation accuracy, and Huang et al. \cite{huang2024robust} introduced RoBox-SAM to enhance robustness under varying box prompt quality. These findings collectively motivate our systematic evaluation of different prompting strategies for fire segmentation, with particular emphasis on bounding box and hybrid approaches (Figure~\ref{fig:bbox_prompt_pipeline}).

\section{Methodology}
\label{sec:method}

\subsection{Prompting Strategy Design}

To evaluate the influence of prompting on fire segmentation performance, we designed five distinct strategies and two hybrid variants. Bounding boxes for all prompt-based strategies are generated using a YOLOv11n detector trained specifically for this work. The detector was trained on the FASSD dataset \cite{wang2022}, the largest publicly available fire detection dataset comprising 95,314 images across general computer vision applications. For our purposes, we selected only images labeled as \emph{fire} and \emph{neither fire nor smoke}, resulting in 51,749 images (12,550 fire and 39,199 neither fire nor smoke). This training setup focuses the model on discriminating fire from non-fire backgrounds, improving recall on fire-related instances. A confidence threshold of 0.3 was applied during inference to prioritize recall and reduce the likelihood of missing potential fire sources. The trained YOLOv11n model achieved the performance metrics shown in Table~\ref{tab:yolo_perf} when evaluated on the FASSD fire-only subset.

\begin{table}[htbp]
\centering
\caption{YOLOv11$n$ Detection Performance on Fire Instances}
\label{tab:yolo_perf}
\scalebox{0.95}{
\begin{tabular}{@{}lcccc@{}}
\toprule
\textbf{Metric} & \textbf{Value} \\
\midrule
Precision (P) & 0.799 \\
Recall (R) & 0.697 \\
mAP@0.5 & 0.797 \\
mAP@0.5:0.95 & 0.520 \\
\bottomrule
\end{tabular}
}
\end{table}

The trained detector serves as the input source for generating bounding box prompts in our segmentation pipeline, ensuring consistent and realistic detection conditions for all prompting strategies:

\begin{enumerate}
    \item \textbf{Automatic (Auto):} SAM2's built-in automatic segmentation without any external prompts, used as a baseline for unprompted performance.
    \item \textbf{Single Positive Point (SP):} One positive point placed at the centroid of each detected bounding box, representing minimal prompt guidance.
    \item \textbf{Single Positive + Single Negative (SP+SN):} Combines a positive centroid point with one negative point located outside all detected boxes, providing explicit separation of fire and background.
    \item \textbf{Multiple Points (MP):} 3x3 positive points grid sampled within each bounding box. Candidate points are filtered using an HSV-based heuristic to retain only those with fire-like color properties.
    \item \textbf{Bounding Box (Box):} Each YOLO-generated bounding box is directly used as a prompt, providing strong spatial priors.
    \item \textbf{Hybrid Variants:}
        \begin{itemize}
            \item \textbf{Box + SP:} Bounding box prompt combined with its centroid point.
            \item \textbf{Box + MP:} Bounding box prompt combined with multiple HSV-filtered positive points.
        \end{itemize}
\end{enumerate}

\section{Experimental Setup}
\label{sec:experiments}

\subsection{Datasets}
We evaluate our approach using three complementary fire datasets that capture diverse scenarios and input modalities. The Khan Fire Segmentation Dataset provides pixel-level annotations of 600 images derived from YouTube videos, covering outdoor fires involving vehicles, buildings, and vegetation under varied lighting and environmental conditions \cite{khan2023efficient}. To assess cross-domain generalization, we incorporate the Roboflow Fire Dataset, a curated collection of 7,040 annotated images from multiple sources, spanning indoor and outdoor environments with diverse fire sizes and appearances \cite{roboflow_fire_2025}. Finally, the Foggia Video Dataset includes five annotated video sequences of fire incidents, enabling evaluation of temporal stability and real-time segmentation performance \cite{foggia_video_fire_2015}.

\subsection{Evaluation Metrics}
Segmentation performance is assessed using five standard metrics. Pixel Accuracy (PA) measures the overall proportion of correctly classified pixels:
\begin{equation}
    \text{PA} = \frac{\sum_{i} n_{ii}}{\sum_{i} t_i},
\end{equation}
where $n_{ii}$ is the number of correctly classified pixels for class $i$ and $t_i$ is the total number of pixels in class $i$. Mean Accuracy (MA) averages per-class accuracy:
\begin{equation}
    \text{MA} = \frac{1}{C} \sum_{i=1}^{C} \frac{n_{ii}}{t_i}.
\end{equation}
Mean Intersection over Union (mIoU) computes the mean overlap ratio:
\begin{equation}
    \text{mIoU} = \frac{1}{C} \sum_{i=1}^{C} 
    \frac{n_{ii}}{t_i + \sum_{j} n_{ji} - n_{ii}},
\end{equation}
while the Frequency Weighted IoU (FWIoU) accounts for class frequency:
\begin{equation}
    \text{FWIoU} = 
    \sum_{i=1}^{C} \frac{t_i}{\sum_k t_k} 
    \frac{n_{ii}}{t_i + \sum_{j} n_{ji} - n_{ii}}.
\end{equation}
We also report the Dice coefficient, equivalent to the F1-score:
\begin{equation}
    \text{Dice} = 
    \frac{2 \sum_i n_{ii}}
         {\sum_i t_i + \sum_i \hat{t}_i},
\end{equation}
where $\hat{t}_i$ is the total number of pixels predicted as class $i$. Additionally, we compute the Mean Absolute Error (MAE) for pixel-wise differences between predicted and ground-truth masks. 

Computational efficiency is evaluated using inference time (ms per frame), frames per second (FPS), memory consumption (MB), and model size (number of parameters).

\subsection{Implementation Details}
Most experiments are conducted on an NVIDIA GeForce GTX 1050 Ti GPU (4GB memory) using PyTorch~2.0 with CUDA~12.4. For computationally intensive models such as PSPNet, SegNet, and FCN-8, we use an NVIDIA RTX~3090 GPU (24GB memory). All models share identical preprocessing pipelines to ensure fair comparison.

\section{Results and Discussion}
\label{sec:results}

\subsection{Overall Performance Analysis}
Table~\ref{tab:khan_results} shows the results for all SAM2.1 variants and prompting strategies gotten from Khan-dataset. Bounding-box prompts outperform point-based and automatic prompts across variants. The Box+MP strategy gives the best overall results, reaching the highest mean IoU (0.643) and Dice (0.755) with SAM2.1 Large. It is also best on mIoU, Dice, MAE, and PA for Tiny and Large, and best on MAE and PA for Small. For Base\_plus, Box alone is slightly better than Box+MP on mIoU and Dice, while automatic prompting performs poorly across board (mIoU $<0.2$), confirming the importance of high-quality prompts.

\begin{table*}[htbp]
\caption{Khan Dataset Results Across SAM2.1 Variants and Prompting Strategies. Best results in \textbf{bold}.}
\label{tab:khan_results}
\centering
\begin{tabular}{|l|cccc|cccc|cccc|cccc|}
\hline
\multirow{2}{*}{\textbf{Strategy}} & \multicolumn{4}{c|}{\textbf{Tiny}} & \multicolumn{4}{c|}{\textbf{Small}} & \multicolumn{4}{c|}{\textbf{Base\_plus}} & \multicolumn{4}{c|}{\textbf{Large}} \\
\cline{2-17}
& \textbf{mIoU} & \textbf{Dice} & \textbf{MAE} & \textbf{PA} & \textbf{mIoU} & \textbf{Dice} & \textbf{MAE} & \textbf{PA} & \textbf{mIoU} & \textbf{Dice} & \textbf{MAE} & \textbf{PA} & \textbf{mIoU} & \textbf{Dice} & \textbf{MAE} & \textbf{PA} \\
\hline
Auto & 0.005 & 0.010 & 2.603 & 0.884 & 0.103 & 0.166 & 139.44 & 0.416 & 0.010 & 0.001 & 0.345 & 0.893 & 0.163 & 0.216 & 17.30 & 0.860 \\
\hline
SP & 0.476 & 0.582 & 18.39 & 0.893 & 0.452 & 0.559 & 18.10 & 0.890 & 0.495 & 0.599 & 15.43 & 0.905 & 0.493 & 0.591 & 18.81 & 0.890 \\
SP+SN & 0.479 & 0.588 & 17.15 & 0.896 & 0.443 & 0.556 & 7.94 & 0.915 & 0.485 & 0.592 & 12.96 & 0.908 & 0.476 & 0.579 & 15.66 & 0.896 \\
MP & 0.557 & 0.671 & 14.25 & 0.927 & 0.560 & 0.671 & 6.74 & 0.945 & 0.587 & 0.697 & 7.43 & 0.946 & 0.591 & 0.700 & 7.42 & 0.945 \\
\hline
Box & 0.629 & 0.744 & 6.63 & 0.953 & \textbf{0.628} & \textbf{0.742} & 4.93 & 0.956 & \textbf{0.627} & \textbf{0.743} & \textbf{6.47} & \textbf{0.954} & 0.640 & 0.752 & 4.65 & 0.959 \\
Box+SP & 0.619 & 0.735 & 7.03 & 0.952 & 0.625 & 0.740 & 4.91 & 0.956 & 0.620 & 0.738 & 7.29 & 0.953 & 0.628 & 0.742 & 5.26 & 0.956 \\
Box+MP & \textbf{0.632} & \textbf{0.745} & \textbf{6.60} & \textbf{0.956} & 0.626 & 0.740 & \textbf{4.34} & \textbf{0.958} & 0.622 & 0.740 & 7.56 & 0.953 & \textbf{0.643} & \textbf{0.755} & \textbf{3.71} & \textbf{0.960} \\
\hline
\end{tabular}
\end{table*}

\subsection{Cross-Dataset Generalization}
On the Roboflow dataset (Table~\ref{tab:roboflow_results}), specialized segmentation networks (FPN, LinkNet) achieved higher overall IoU and FWIoU, but SAM2.1 models remain competitive despite being trained without domain-specific fine-tuning. This difference is expected because FPN and LinkNet are optimized in a fully supervised manner on fire-specific data, giving them an inherent advantage in this narrow domain. By contrast, SAM and its variants follow a promptable, zero-shot paradigm trained on large and diverse datasets without task-specific supervision. While this often results in lower accuracy compared to specialized models, the key strength of SAM lies in its ability to generalize across domains and tasks without retraining. Mobile variants (TinySAM, MobileSAM) further highlight this flexibility by offering competitive performance while remaining suitable for edge deployment.

\begin{table}[htbp]
\caption{Roboflow Dataset Results (Segmentation Models vs. SAM and SAM2.1 Variants). Best results in \textbf{bold}.}
\label{tab:roboflow_results}
\centering
\scalebox{0.85}{
\begin{tabular}{|l|c|c|c|c|}
\hline
\textbf{Model} & \textbf{PA} & \textbf{MA} & \textbf{mIoU} & \textbf{FWIoU} \\
\hline
FPN (MobileNetV2) & \textbf{0.991} & \textbf{0.930} & \textbf{0.883} & \textbf{0.983} \\
LinkNet (MobileNetV2) & 0.991 & 0.928 & 0.881 & 0.982 \\
\hline
PSPNet & 0.982 & 0.789 & 0.758 & 0.965 \\
SegNet & 0.983 & 0.781 & 0.763 & 0.966 \\
FCN-8 & 0.985 & 0.832 & 0.799 & 0.971 \\
\hline
SAM2.1 Large + Box & 0.987 & 0.858 & 0.667 & 0.981 \\
SAM2.1 Base\_plus + Box+MP & 0.986 & 0.876 & 0.663 & 0.978 \\
TinySAM + Box & 0.985 & 0.835 & 0.619 & 0.977 \\
MobileSAM + Box & 0.987 & 0.871 & 0.659 & 0.979 \\
\hline
\end{tabular}
}
\end{table}

\subsection{Video Segmentation and Computational Efficiency}
We evaluated multiple segmentation model variants on five fire videos (\textit{Fire1, Fire3, Fire4, Fire8, Fire13}) from the Foggia dataset to assess their suitability for deployment. Each configuration combines YOLO-based detection for bounding box generation (Box) with different segmentation backbones: SAM2.1 Large, MP+SAM2.1 Base\_plus, TinySAM, and MobileSAM. The bounding box step substantially reduces the segmentation area, improving inference stability and efficiency by restricting computation to regions of interest.

Table~\ref{tab:fire_results} presents detailed results, including per-video inference time, frames per second (FPS), peak GPU memory consumption, and overall model size. 

\begin{table*}[htbp]
\caption{Computational efficiency results for all fire videos with bounding box-based segmentation. Best results in \textbf{bold}.}
\label{tab:fire_results}
\centering
\begin{tabular}{|l|c|c|c|c|c|c|c|c|c|}
\hline
\textbf{Model} & \textbf{Size (MB)} & \textbf{Fire1 (ms)} & \textbf{Fire3 (ms)} & \textbf{Fire4 (ms)} & \textbf{Fire8 (ms)} & \textbf{Fire13 (ms)} & \textbf{Avg (ms)} & \textbf{FPS} & \textbf{Peak Mem (MB)} \\
\hline
Box + SAM2.1 Large         & 866.08 & 2031.32 & 1908.51 & 2126.35 & 2113.36 & 2058.86 & 2047.68 & 0.47 & 2522.64 \\
Box + MP+SAM2.1 Base\_plus & 318.30 & 915.24  & 922.68  & 812.22  & 1116.17 & 884.69  & 930.60  & 1.09 & 1884.20 \\
Box + TinySAM              & 48.52  & \textbf{241.53}  & 288.78  & 241.55  & \textbf{335.56}  & 227.82  & 267.05  & 3.82 & \textbf{332.18} \\
Box + MobileSAM            & \textbf{48.52}  & 250.53  & \textbf{269.42}  & \textbf{227.05}  & 361.95  & \textbf{206.98}  & \textbf{263.99}  & \textbf{4.83} & 346.46 \\
\hline
\end{tabular}
\end{table*}

Among the tested models, MobileSAM with Box achieved the highest efficiency at $4.83$ FPS with $346.46$ MB peak memory usage, followed closely by TinySAM at $3.82$ FPS and a slightly smaller memory footprint. Both are an order of magnitude faster and lighter than SAM2.1 Large, which operated at only $0.47$ FPS and consumed over $2.5$ GB of memory. The MP+SAM2.1 Base\_plus configuration offered an intermediate trade-off between speed and accuracy, but still fell behind TinySAM and MobileSAM in efficiency. 

None of the evaluated variants achieved real-time video segmentation (commonly defined as $\geq$21 FPS). However, the integration of bounding box pre-processing significantly reduced computational complexity across all models, making lightweight variants feasible for latency-tolerant scenarios such as periodic monitoring, post-event analysis, or batch processing on edge devices. In general, lightweight variants such as MobileSAM and TinySAM demonstrate efficiency characteristics that make them potentially suitable for deployment on resource-constrained or latency-tolerant platforms, whereas the MP+SAM2.1 Base plus  configurations offer a middle ground for systems that prioritize accuracy over speed. In contrast, SAM2.1 Large remains more appropriate for centralized, high-power environments where segmentation quality is prioritized over computational efficiency.

\begin{figure}[htbp]
    \centering
    \includegraphics[width=0.95\linewidth]{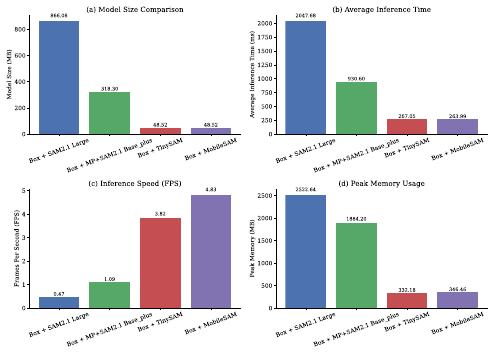}
    \caption{Comparison of computational efficiency metrics for all fire videos using bounding box-based segmentation.
    Metrics include (a) model size, (b) average inference time, (c) inference speed (FPS), and (d) peak memory usage for different models.}
    \label{fig:fire_metrics}
\end{figure}

Overall, these results highlight a flexible trade-off between speed, memory, and deployment setting, while underscoring the need for further optimization to approach true real-time performance (Figure~\ref{fig:fire_metrics}).  

\section{Discussion}
The results across datasets suggest three consistent observations. First, prompt quality is the dominant driver of segmentation accuracy as SAM variants with the Yolov11n bounding box prompts outperform variants with point-only and automatic prompt strategies, with the Box+MP hybrid yielding the most reliable masks (Table~\ref{tab:khan_results}). These gains are prominent when bounding box is used as prompt to SAM2.1, which indicates that improvements primarily stem from stronger spatial priors rather than model size alone. 

Second, on the Roboflow dataset (Table~\ref{tab:roboflow_results}), specialized networks (FPN and LinkNet) achieve higher mean IoU and FWIoU, which align with their fully supervised, task-specific training; by contrast, SAM variants, though promptable and zero-shot, remain competitive without domain-specific fine-tuning.

Third, for video inference, the bounding box stage reduces the processed area and improves efficiency (Table~\ref{tab:fire_results}, Fig.~\ref{fig:fire_metrics}). Additionally, lightweight variants (MobileSAM, TinySAM) provide the highest throughput and lowest memory usage among the evaluated configurations; however, none of the models reach a typical real-time threshold (e.g., $\geq$21 FPS). This suggests current feasibility for latency-tolerant settings while indicating a need for further optimization to close the real-time gap.

\section{Conclusion}
This work presents a comprehensive evaluation of SAM2.1 and SAM mobile variants for fire segmentation, with a focus on bounding box prompting. Across multiple datasets, the results indicate that bounding box–assisted prompting consistently improves segmentation metrics over point-based or automatic strategies. Mobile-oriented variants (e.g., MobileSAM, TinySAM) exhibit favorable efficiency on the tested hardware; for example, MobileSAM with bounding box prompting attains 4.83 FPS with a peak memory usage of approximately 346~MB. None of the evaluated configurations achieve typical real-time throughput on the tested setup, highlighting the need for further optimization to close the performance gap. Overall, our study establishes a systematic benchmark of SAM2-based fire segmentation, highlighting the trade-offs between accuracy and efficiency across prompting strategies and model variants. The findings offer a foundation for future work on adapting promptable segmentation models to fire monitoring tasks under diverse computational and operational constraints.


\begin{thebibliography}{00}

\bibitem{ahrens2010smoke} M. Ahrens, "Smoke alarm presence and performance in U.S. home fires," \emph{Fire Technology}, vol. 47, no. 3, pp. 699--720, 2010.

\bibitem{khan2023efficient} K. Muhammad, H. Ullah, S. Khan, M. Hijji, and J. Lloret, "Efficient fire segmentation for internet-of-things-assisted intelligent transportation systems," \emph{IEEE Trans. Intell. Transport. Syst.}, vol. 24, no. 11, pp. 13141--13150, 2023.

\bibitem{muhammad2023efficient} K. Muhammad et al., "Efficient deep CNN-based fire detection and localization in video surveillance applications," \emph{IEEE Trans. Syst. Man Cybern. Syst.}, vol. 49, no. 7, pp. 1419--1434, 2019.

\bibitem{kirillov2023segment} A. Kirillov et al., "Segment anything," in \emph{Proc. IEEE/CVF Int. Conf. Comput. Vis.}, 2023, pp. 4015--4026.

\bibitem{ravi2024sam2} N. Ravi et al., "SAM 2: Segment anything in images and videos," arXiv preprint arXiv:2408.00714, 2024.

\bibitem{zhangunleashing} Y. Zhang and Z. Shen, "Unleashing the potential of SAM2 for biomedical images and videos: A survey,"  arXiv preprint arXiv.org, https://arxiv.org/abs/2408.12889.

\bibitem{yurobotics} J. Yu et al., "Sam 2 in robotic surgery: An empirical evaluation for robustness and generalization in surgical video segmentation,"  arXiv preprint arXiv.org, http://dx.doi.org/10.48550/arXiv.2408.04593.

\bibitem{senguptamedical} S. Sengupta, S. Chakrabarty, and R. Soni, "Is Sam 2 better than Sam in medical image segmentation?,"  arXiv preprint arXiv.org, https://arxiv.org/abs/2408.04212. 

\bibitem{lian2024evaluation} S. Lian and H. Li, "Evaluation of segment anything model 2 in the underwater environment," arXiv preprint arXiv:2408.02924, 2024.

\bibitem{tang2024evaluating} L. Tang and B. Li, "Evaluating SAM2's role in camouflaged object detection: From SAM to SAM2," arXiv preprint arXiv:2407.21596, 2024.

\bibitem{marinaccio2025seeing} M. Marinaccio and F. Afghah, "Seeing heat with color: RGB-only wildfire temperature inference from SAM-guided multimodal distillation using radiometric ground truth," arXiv preprint arXiv:2505.01638, 2025.

\bibitem{pesonen2025detecting} J. Pesonen et al., "Detecting wildfires on UAVs with real-time segmentation trained by larger teacher models," in \emph{Proc. IEEE/CVF Winter Conf. Appl. Comput. Vis.}, 2025, pp. 1--10.

\bibitem{sun2024efficient} X. Sun et al., "On efficient variants of segment anything model: A survey," arXiv preprint arXiv:2410.04960, 2024.

\bibitem{thanammal2014effective} K. K. Thanammal, J. S. Jayasudha, R. R. Vijayalakshmi, and S. Arumugaperumal, 
"Effective Histogram Thresholding Techniques for Natural Images Using Segmentation," \emph{Journal of Image and Graphics}, vol. 2, no. 2, pp. 113--116, Dec. 2014. 
https://doi.org/10.12720/joig.2.2.113-116.

\bibitem{kaur2014fcm} R. Kaur and G. Malik, 
"An Image Segmentation Using Improved FCM Watershed Algorithm and DBMF," \emph{Journal of Image and Graphics}, vol. 2, no. 2, pp. 106--112, Dec. 2014. 
https://doi.org/10.12720/joig.2.2.106-112.

\bibitem{richard2013fuzzy} N. Richard, C. Fernandez-Maloigne, C. Bonanomi, and A. Rizzi, 
"Fuzzy Color Image Segmentation using Watershed Transform," \emph{Journal of Image and Graphics}, vol. 1, no. 3, pp. 157--160, Sept. 2013. 
https://doi.org/10.12720/joig.1.3.157-160.

\bibitem{toulouse2015automatic} T. Toulouse, L. Rossi, T. Celik, and M. Akhloufi, 
"Automatic fire pixel detection using image processing: A comparative analysis of rule-based and machine learning based methods," 
\emph{Signal Image Video Process.}, vol. 10, no. 4, pp. 647--654, 2016.

\bibitem{sowahfuzzy} R. Sowah et al, 
"Design and implementation of a fire detection and control system for automobiles using fuzzy logic," 
\emph{IEEE Industry Applications Society Annual Meeting}, https://doi.org/10.1109/ ias.2016.7731880, Oct. 2016.

\bibitem{sowah2018neurofuzzy}
R. Sowah, A. Ofoli, K. Koumadi, G. Osae, G. Nortey, A. M. Bempong, B. Agyarkwa, and K. O. Apeadu,
"Design and implementation of a fire detection and control system with enhanced security and safety for automobiles using neuro-fuzzy logic,"
\emph{Proc. IEEE 7th Int. Conf. Adaptive Sci. \& Technol. (ICAST)}, pp. 1--8, 2018.

\bibitem{khan2024yolo} N. Jegham et al., 
"YOLO Evolution: A Comprehensive Benchmark and Architectural Review of YOLOv12, YOLO11, and Their Previous Versions," 
arXiv preprint arXiv:2411.00201, 2024.

\bibitem{liusafe2025}  S. Liu, Y. Xue et al. "Segmentation of Any Fire Event (SAFE): A Rapid and High-Precision Approach for Burned Area Extraction Using Sentinel-2 Imagery ," https://doi.org/10.3390/rs17010054, 2025.

\bibitem{zhao2023mobilesam} C. Zhao et al., 
"MobileSAM: Making SAM mobile with decoupled distillation," 
arXiv preprint arXiv:2312.17481, 2023.

\bibitem{shu2024tinysam} H. Shu et al., 
"TinySAM: Pushing the envelope for efficient segment anything model," 
arXiv preprint arXiv:2312.13789, 2024.

\bibitem{nanosam} NVIDIA AI IOT, "NanoSAM" [Online]. Available: https://github.com/NVIDIA-AI-IOT/nanosam

\bibitem{bonazzi2025picosam2} P. Bonazzi et al., 
"PicoSAM2: Low-latency segmentation in-sensor for edge vision applications," 
arXiv preprint arXiv:2506.18807, 2025.

\bibitem{rafaeli2024prompt} O. Rafaeli et al., 
"Prompt-based segmentation at multiple resolutions and lighting conditions using segment anything model 2," 
arXiv preprint arXiv:2408.06970, 2024.

\bibitem{pei2024evaluation} J. Pei, Z. Zhou, and T. Zhang, 
"Evaluation study on SAM 2 for class-agnostic instance-level segmentation," 
arXiv preprint arXiv:2409.02567, 2024.

\bibitem{li2024amsam} Y. Li et al., 
"AM-SAM: Automated prompting and mask calibration for segment anything model," 
arXiv preprint arXiv:2410.09714, 2024.

\bibitem{huang2024robust} Y. Huang et al., 
"Robust box prompt based SAM for medical image segmentation," 
in \emph{Proc. Int. Workshop Mach. Learn. Med. Imaging}, 2024, pp. 1--11.


\bibitem{wang2022}  M. Wang, L. Jiang, P. Yue, D. Yu, and T. Tuo. "FASDD: An open-access 100,000-level flame Detection dataset for deep learning in fire detection ," doi:10.5194/essd-2022-394, Nov. 2022.


\bibitem{roboflow_fire_2025} Roboflow, "Fire Segmentation Part 1 Dataset," 
Available: \url{https://universe.roboflow.com/firesegpart1/fire-seg-part1/dataset/21}

\bibitem{foggia_video_fire_2015} P. Foggia et al., "Real-time fire detection for video-surveillance applications using a combination of experts based on color, shape, and motion," \textit{Computer Vision and Image Understanding}, vol. 143, pp. 144-154, 2015.

\end{thebibliography}
\end{document}